\documentclass[11pt]{article}

\usepackage[
  shownumpages,          
  citingstyle=numbers,     
  bibliostyle=plainnat,    
  bibfile=refs,      
]{brainlab}

\usepackage{shortcuts}

\usepackage{lipsum}

\setbrainmeta{
  title={Extreme Low-Bit Inference in Reasoning Models: \\Failure Modes and Targeted Recovery},
  authors={Ekaterina Alimaskina, Darya Rudas, Denis Shveykin, Gleb Molodtsov, Pavel Vasiliev, \\Aleksandr Beznosikov
  },
  affiliations={
    BRAIn Lab, {\scriptsize{Moscow, Russia}} \vspace{2mm}
  }, 
  abstract={
Large Reasoning Models (LRMs) rely on long reasoning traces, making inference expensive. While low-bit quantization reduces per-token decoding cost, we show that aggressive 2-bit inference can fail to deliver end-to-end speedup because instability in the generation process inflates total token count. Instead of merely lowering answer accuracy, 2-bit quantization often produces much longer traces with repetitive loops, budget exhaustion, delayed commitment, and unclosed reasoning segments. We analyze full reasoning traces of Qwen3 reasoning models across mathematical and commonsense benchmarks and show that accuracy degradation is tightly linked to these process-level failures. To address them, we introduce two lightweight controls: FP16 planning, which gives the 2-bit model a short high-precision outline, and loop rescue, which detects repetitive traces and either commits to an earlier answer or falls back to FP16. On MATH-500, loop rescue improves Qwen3-8B accuracy from 17.2\% to 74.2\%, while planning plus loop rescue improves Qwen3-32B from 65.0\% to 87.2\%. Overall, our results show that extreme low-bit reasoning becomes practical when its failures are treated as controllable generation pathologies: with lightweight detection and selective FP16 support, 2-bit inference can recover accuracy while preserving real end-to-end speed.
Our code is available at: \url{https://github.com/brain-lab-research/quantized-reasoning}.
  },
}

\begin{document}
\begin{mainpart}

\allowdisplaybreaks
\vspace{-8mm}
\section{Introduction}
\label{sec:intro}
Large Reasoning Models (LRMs) have recently emerged as a major direction in the development of Large Language Models (LLMs).
Their strong performance on complex tasks is largely enabled by long reasoning traces, which allow the model to search for a solution, verify intermediate steps, and refine its conclusions before producing the final answer.
This mechanism, however, substantially increases inference cost, as higher reasoning quality often  requires much longer outputs.
As a result, even small reductions in the cost of each decoding step can lead to notable savings.

\begin{wrapfigure}{r}{0.4\textwidth}
    \vspace{-6mm} 
    \centering
    \includegraphics[width=0.4\textwidth]{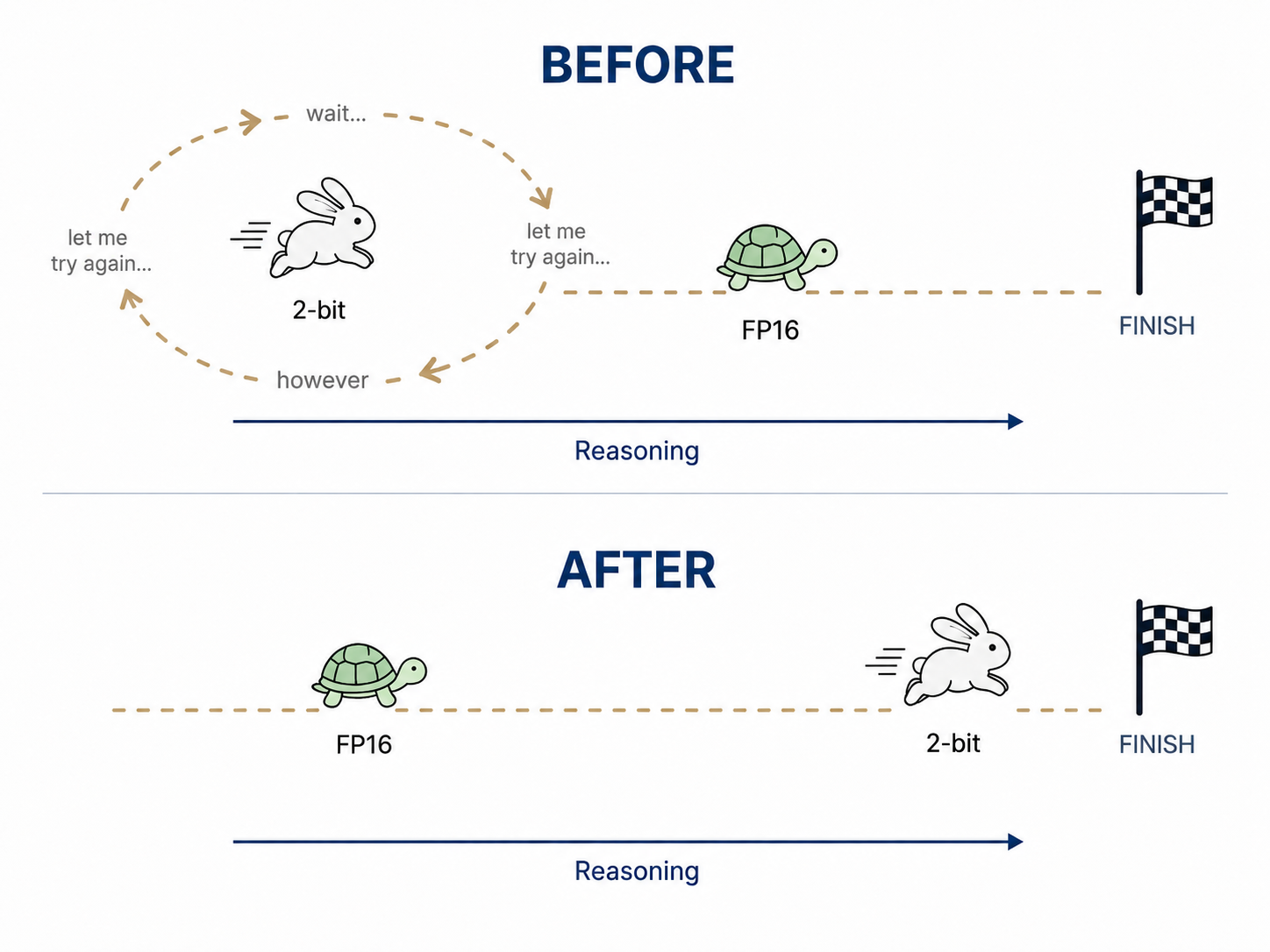}
    \vspace{-8mm}
    \caption{High-level view of low-bit reasoning.}
    \label{fig:teaser}
    \vspace{-4mm}
\end{wrapfigure}

Quantization is one of the main tools for accelerating LLM inference \citep{gholami2022survey}.
In non-reasoning settings, modern quantization methods preserve quality well even under substantial compression.
For LRMs, the situation is more subtle. Unlike standard generation, small perturbations introduced by low-bit inference can accumulate over long reasoning trajectories and ultimately affect the final answer \citep{quantization_meets_reasoning,when_reasoning_meets_compression}. Yet, despite its practical importance, this regime remains only weakly understood. Many existing evaluations conflate quantization effects with prompt formatting, decoding choices, and budget selection, obscuring whether degradation is truly quantization-induced.

To close this gap, we ask a more direct question: why does low-bit reasoning fail, and how to fix it? Instead of reducing the problem to the final answer accuracy alone, we analyze the full reasoning trajectory to identify the specific behaviors through which quantization degrades performance.
We find that these failures cannot be explained simply by weaker task-solving ability, but instead arise from several distinct mechanisms. In particular, the quantized model often derives the correct answer within its trace yet fails to stop and commit to it. Figure~\ref{fig:teaser} shows the intuition: 2-bit reasoning is cheap per token, but demands reliable answer extraction.
Based on this diagnosis, we show that targeted interventions in the identified failure stages substantially close the gap to full-precision reasoning.

\section{Related Work}
\label{sec:related}
\paragraph{Post-Training Quantization (PTQ).}
Quantization is one of the main compression techniques for LLMs, which reduces the numerical precision of model parameters by representing them with lower-bit formats.
A common approach is PTQ, which is attractive because it can be applied to an already trained model without retraining or task-specific fine-tuning \citep{gptq,smoothquant,awq,gsq,harp}.
4-bit quantization has already become a relatively well-studied setting which typically leads to only limited degradation compared to full precision \citep{dettmers2023case}.
To further improve inference efficiency, recent work investigated more aggressive regimes, such as 2--3-bits, which tend to be considerably more fragile \citep{lowbit_survey, fast_2bit_inference, abq_llm,when_reasoning_meets_compression}.
However, quantization for reasoning generation remains underexplored. Motivated by its importance, we focus on low-bit quantization as a way to speed-up inference of reasoning models.

\paragraph{Quantization and reasoning degradation.}
Prior work suggested that quantization affects reasoning not only as a uniform quality drop, but also by changing the reasoning process itself.
\citet{quantization_meets_reasoning} suggested that low-bit models may preserve the high-level solution idea while becoming more vulnerable to local mistakes.
\citet{quantization_hurts_reasoning} further demonstrated that degradation strongly depends on the evaluation regime, including model architecture, and both calibration and evaluation data.
They also found that quantized models benefit less from extended reasoning than full-precision ones, suggesting an early switching to full-precision completion.
Despite \citet{quantlrm} showing that W4A16 preserves reasoning quality, going beyond 4 bits remained fragile while offering additional speed-up.

\paragraph{Quantized and speculative inference.}
The idea of utilizing low-bit inference benefits while taking into account potential quality drops motivated verification-based setups similar to speculative decoding. Here a low-cost model proposes candidate tokens and a more accurate model verifies them.
\citet{qspec} suggested using low-precision weight-activation formats in the draft model.
\citet{mlspecqd} further lowered draft-stage cost through an additional speculative level with a quantized auxiliary model.
\citet{speq} avoided a separate drafter by constructing a low-bit draft model from the target model's own weights.
\citet{quasar} introduced a quantized verification stage to accelerate the parallel verification.
However, these methods operate on short chains, rather than letting the low-bit model perform longer reasoning independently.
This could offer additional speedups, but requires understanding which stages remain reliable under aggressive quantization and which are the most susceptible.

\paragraph{Collaborative efficient reasoning.}
A related line of work accelerates reasoning by routing different parts of the reasoning process to models of different capacity.
\citet{cope, tandem} showed that smaller executors can benefit from compact guidance produced by stronger models: initial thinking can improve performance across subjects, while explicit plans provide a lightweight interface for transferring high-level foresight.
\citet{splitreason} and \citet{rsd} introduced adaptive intervention, where control is transferred to a stronger model only at difficult reasoning steps, as determined by learned switching or reward signals.
\citet{scot} and \citet{specreason} moved speculation to reasoning-level units, such as candidate chains of thought or individual reasoning steps.
However, large/small collaboration introduces fundamental mismatch in model capacity, training data, reasoning style, and output format.
This leaves a gap for a complementary direction: pairing a reasoning model with its quantized counterpart, which requires identifying when quantized reasoning is reliable and when full-precision is needed.

\paragraph{Overthinking.}
Another challenge in efficient reasoning is overthinking.
Reasoning models may continue to verify, revise, or repeat intermediate conclusions even after a plausible or correct answer has already appeared.
This increases inference cost and makes latency less predictable.
For example, \citet{terminator} showed that the final answer can appear substantially before the end of the generated chain-of-thought, with a spike in token confidence and log-probability around this first-answer moment.
This suggests that later reasoning may often reflect verification, rewriting, or formatting rather than genuine solution search.
At the same time, existing work uses different experimental setups and token budgets, making reported speedups and accuracy trade-offs difficult to compare directly \citep{splitreason,speculative_thinking,scot,specreason}.

Therefore, the main question is not only whether quantization can make reasoning faster, but also how it should be used.
We need to understand what parts of the reasoning process can be safely delegated to a low-bit model, where low-bit degradation first appears, and how the inference setup depends on quantization aggressiveness, context length, task difficulty, and the required quality--latency trade-off.
This motivates our study of quantized reasoning as a phase-dependent process rather than a uniform replacement of the full-precision model with a low-bit one.

\paragraph{Contributions.} Our contributions are as follows:
\begin{itemize}[nosep, leftmargin=*]
\item 
We show that extreme low-bit quantization qualitatively degrades LRM reasoning traces, inducing looping, delayed commitment, and budget exhaustion. We identify two dominant failure modes: path-finding failure (no parseable answer is ever produced) and commitment failure (an answer is found but reasoning never terminates) -- and use them for targeted mitigation.

\item 
We introduce a regime-based taxonomy classifying model--benchmark pairs into four regimes by accuracy delta and think-closed rate, and show it predicts when selective FP16 planning and loop-aware control suffice versus when full-precision fallback is required.

\item
We propose a lightweight inference pipeline for 2-bit LRMs that uses FP16 for critical planning steps, detects and breaks reasoning loops, and selectively falls back to full precision. This recovers most of the FP16 accuracy gap while preserving the latency benefits of quantization.
\end{itemize}

\section{Setup}
\label{sec:setup}
We study quantization in a controlled setting using Qwen-family reasoning models~\citep{qwen3} at two representative scales, 8B and 32B. We focus on quantization schemes that are both widely used in prior work and practical in existing inference stacks. In particular, we compare the \texttt{FP16} baseline with GPTQ-style weight-only quantization~\citep{gptq} in \texttt{W4A16} and \texttt{W2A16}. This choice lets us compare a near-lossless regime against a much more aggressive compression setting, while keeping the evaluation setup standard.

Our benchmark suite covers both mathematical and commonsense reasoning: AIME 2026~\citep{aime2026}, GPQA-Diamond~\citep{rein2023gpqa}, MATH-500~\citep{hendrycks2021math}, GSM8K~\citep{cobbe2021gsm8k}, StrategyQA~\citep{geva2021strategyqa}, WinoGrande~\citep{sakaguchi2020winogrande}, ARC-Easy / ARC-Challenge~\citep{clark2018arc}, and PIQA~\citep{bisk2020piqa}. Unless stated otherwise, all experiments in the main part are run with the full 32K reasoning budget; shorter budgets are analyzed separately in Section \ref{sec:budget}. We use a temperature of 0.6, as recommended by the Qwen authors; a separate ablation study on temperature is provided later. We use the same evaluation protocol across all setups so that changes in quality and latency can be attributed to precision rather than prompt or decoding differences.

\section{Looking Inside the Reasoning Trace}
\label{sec:trace_analysis}

Most quantization evaluations summarize model quality with final-answer accuracy. This is a necessary metric, but for reasoning models it is incomplete: a reasoning trace is not only an answer, but a process. The model searches for a solution path, verifies intermediate steps, revises its reasoning, and eventually commits to a final answer.
We therefore evaluate both the final answer and the evolution of the reasoning trace, allowing us to separate changes to the answer from changes to the internal reasoning dynamics.

\subsection{Trace-level metrics}

For each generation, we track a small set of signals that describe how the reasoning trace evolves before the final answer is parsed:
 
\begin{table}[htbp]
\centering
\small
\begin{minipage}{0.54\columnwidth}
\raggedright
\begin{itemize}[leftmargin=*]
    \item \textbf{Reasoning length}: number of generated reasoning tokens.
    \item \textbf{Hit-limit rate}: fraction of traces reaching the token budget.
    \item \textbf{Think-closed rate}: fraction of traces closing with \texttt{</think>}.
    \item \textbf{TTFA (time to first answer)}: token position of the first parseable answer.
    \item \textbf{Loop rate}: fraction of traces with repeated local $n$-gram patterns.
\end{itemize}
\end{minipage}%
\hfill
\begin{minipage}{0.44\columnwidth}
\vspace{-6mm}
\centering
\caption{
Correlation between accuracy and trace-metric changes when moving from FP16 to 2-bit quantization.
}
\label{tab:trace_metric_correlations}
\vspace{-2mm}
\begin{tabular}{lrr}
\toprule
$\Delta$ metric & Pearson $r$ & Spearman $\rho$ \\
\midrule
Loop rate & $-0.85$ & $-0.88$ \\
Hit max & $-0.83$ & $-0.89$ \\
Reasoning length & $-0.73$ & $-0.73$ \\
Think closed & $+0.73$ & $+0.90$ \\
TTFA & $+0.70$ & $+0.62$ \\
\bottomrule
\end{tabular}
\end{minipage}
\vspace{-4mm}
\end{table}

Together, these metrics complement accuracy by describing the reasoning trace before the final answer is parsed: its length, termination behavior, first answer position, and repetitive patterns.

\subsection{Quantization changes the trace dynamics}

\begin{wrapfigure}{r}{0.5\textwidth}
\vspace{-12mm}
\centering
\includegraphics[width=0.48\textwidth]{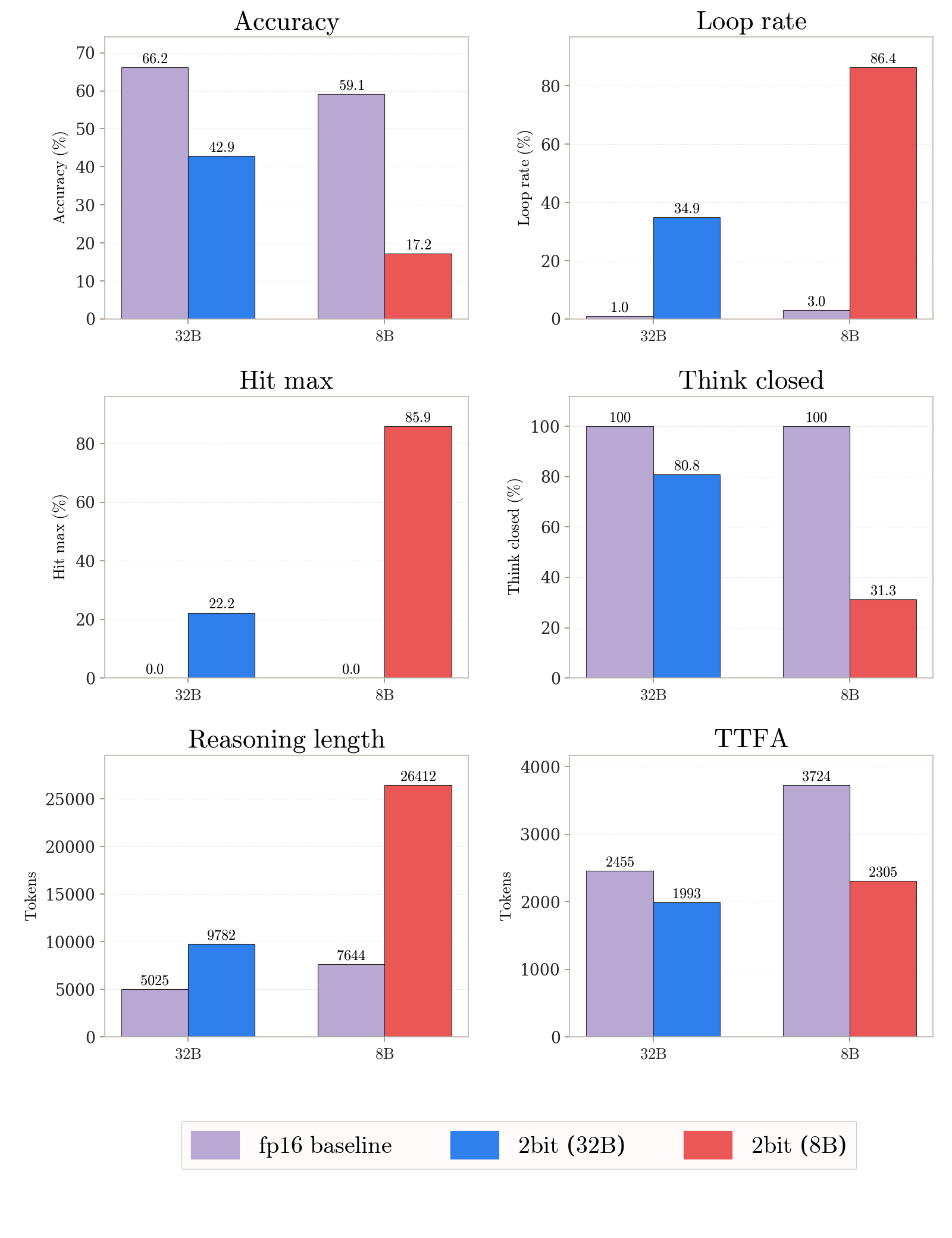}
\vspace{-6mm}
\caption{
Trace-level changes under 2-bit quantization on GPQA-Diamond. FP16 and 2-bit runs for Qwen3-32B and Qwen3-8B across accuracy, loop rate, hit-limit, think-closed rate, reasoning length \& TTFA.
}
\label{fig:gpqa_trace_metrics}
\vspace{-14mm}
\end{wrapfigure}

We first examine whether the accuracy loss under 2-bit quantization is reflected inside the reasoning trace. Figure~\ref{fig:gpqa_trace_metrics} shows a representative comparison between FP16 and 2-bit inference for both models.

The figure shows that quantization affects not only the final answer, but also the trajectory that produces it. Under 2-bit inference, accuracy drops while traces become less stable: loops and budget exhaustion increase, and the model closes the reasoning segment less reliably. This effect is mild for Qwen3-32B but much stronger for Qwen3-8B.

This pattern is consistent across datasets and models. Larger accuracy drops are associated with stronger trace-level degradation. Table~\ref{tab:trace_metric_correlations} summarizes this by correlating the change in accuracy with the change in each trace metric when moving from FP16 to 2-bit.

The correlations show that model–benchmark pairs with larger accuracy drops also tend to produce longer traces, more loops, budget exhaustion, and fewer closed reasoning segments. Thus, 2-bit quantization often changes the dynamics of generation itself, not only the final answer.

\subsection{From repeated reasoning to failure modes}

Longer traces are not necessarily worse: difficult problems may require more tokens. However, under 2-bit quantization, longer traces often coincide with lower accuracy, lower think-closed rate, and higher loop rate. This suggests that many extra tokens are not productive, but repeated reasoning.

\begin{wrapfigure}{r}{0.65\textwidth}
\vspace{-2mm}
\centering
\includegraphics[width=0.65\textwidth]{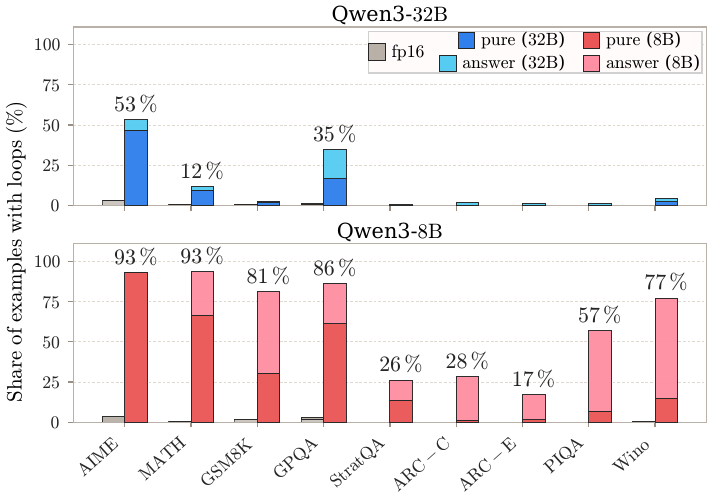}
\vspace{-6mm}
\caption{
Pure and answer loops under 2-bit quantization. Looped traces are split by whether a parseable answer appeared before the loop.
}
\label{fig:pure_answer_loops}
\vspace{-8mm}
\end{wrapfigure}

This repetition points to 2 failures:
\begin{itemize}[nosep,leftmargin=*]
    \item \textbf{Path-finding failure.} The model does not reach a parseable answer (i.e., in \texttt{\textbackslash boxed\{\}}) before the trace becomes repetitive. Instead, it keeps exploring an unproductive path.

    \item \textbf{Commitment failure.} The model reaches a parseable answer, but does not stop. It continues checking, revising, or restating the solution instead of committing to the answer.
\end{itemize}

This connects to a broader phenomenon in reasoning models: long CoT traces can contain redundant verification and delayed stopping \citep{sui2025stopoverthinking,yang2025deer}. Prior work mainly treats this as an efficiency issue; we show that extreme quantization couples it with accuracy degradation.

In practice, both path-finding and commitment failures often manifest as loops. We therefore split looped traces by whether a parseable answer appeared before the loop: loops before an answer indicate failure to reach a candidate solution, while loops after an answer indicate failure to commit.

Figure~\ref{fig:pure_answer_loops} shows that both failure modes are amplified under 2-bit quantization. For Qwen3-32B, loops remain concentrated on harder reasoning tasks such as AIME, MATH, and GPQA. For Qwen3-8B, the effect is broader: loops appear across almost all benchmarks, with large rates on MATH, GPQA, GSM8K, PIQA, and WinoGrande.

\begin{tcolorbox}[
    colback=takeawaybg,
    colframe=takeawayborder,
    boxrule=0.5pt,
    arc=2pt,
    left=7pt,
    right=7pt,
    top=6pt,
    bottom=6pt,
    boxsep=0pt,
    enhanced
]
\textbf{Takeaway.} Accuracy shows that 2-bit quantization hurts reasoning; trace metrics show how reasoning changes. Loops are one diagnostic signal: before an answer they suggest path-finding failure, after an answer they suggest failure to commit.
\end{tcolorbox}

\section{Generation Budget Changes the Observed Failure Mode}
\label{sec:budget}

\begin{wrapfigure}{l}{0.42\textwidth}
\vspace{-5mm}
\centering
\includegraphics[width=0.42\textwidth]{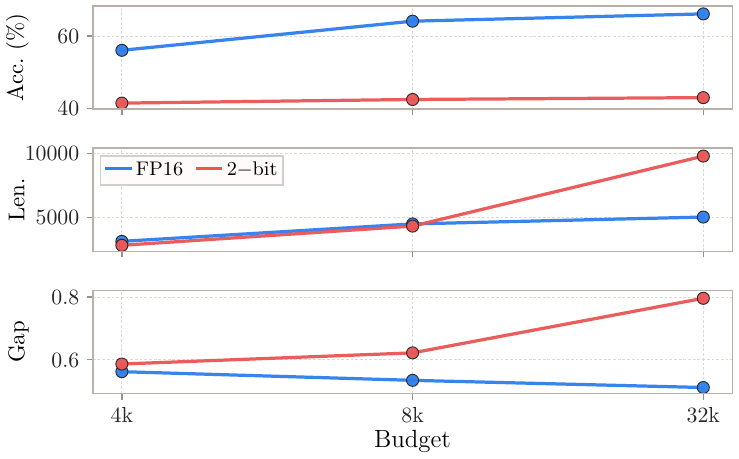}
\vspace{-4mm}
\caption{
Budget dependence on Qwen3-32B for GPQA-Diamond. We compare FP16 and 2-bit runs under 4k, 8k, and 32k generation limits. 
}
\label{fig:budget_gpqa}
\vspace{-6mm}
\end{wrapfigure}

The trace-level failures above directly affect latency. Per-token speedup is a poor proxy for end-to-end efficiency: a 2-bit model that repeats, reverifies, or exhausts its budget can easily erase its throughput advantage. Reasoning speed is jointly determined by token cost and generation length, quantization degrades the latter.

This interaction is easy to miss. If the generation budget is short, repeated reasoning is truncated early: the model has less opportunity to enter long loops, and the resulting trace may look artificially bounded. With a longer budget, the same model can continue generating, exposing heavy-tail behavior in reasoning length, delayed commitment, and budget exhaustion. Therefore, the choice of maximum generation length is important.

Evaluation budgets vary widely across acceleration work, with several recent systems capping generation at around 8k tokens \citep{tandem,speccot}. This covers typical reasoning traces but misses the long-context regime -- up to 16k–32k tokens -- that modern reasoning models are designed to exploit. We therefore use 32k tokens as our main setting.
To quantify the effect of budget, we evaluate the same model--dataset pairs under 4k, 8k, and 32k generation limits. 

We track accuracy, reasoning length, and the normalized commit gap: the fraction of the trace generated after the model first produces a parseable answer.
A small gap means that the model produces an answer near the end of the trace and stops soon after. A large gap means that the model reaches a candidate answer early, but continues reasoning for a substantial part of the generation.

Figure~\ref{fig:budget_gpqa} illustrates the budget dependence. Increasing the budget helps the FP16 model: accuracy improves from the 4k to the 32k setting, while the normalized commit gap decreases. In contrast, the 2-bit model does not convert the extra budget into accuracy. Its accuracy remains nearly flat, but its reasoning length grows substantially, and the normalized commit gap increases sharply.

This shows why short-budget evaluations can systematically overestimate the end-to-end benefit of low-bit reasoning. At 4k or 8k, repeated reasoning is truncated early: the model has less room to enter long loops or to continue after the first candidate answer. This hides part of the cost introduced by process-level failures, so per-token speedups translate too optimistically into per-example speedups. At 32k, the same model can reveal a different behavior: it spends many more tokens without a corresponding quality gain. In this regime, quantization changes not only the cost per generated token, but also the number of tokens generated before termination.

\begin{tcolorbox}[
    colback=takeawaybg,
    colframe=takeawayborder,
    boxrule=0.5pt,
    arc=2pt,
    left=7pt,
    right=7pt,
    top=6pt,
    bottom=6pt,
    boxsep=0pt,
    enhanced
]
\textbf{Takeaway.} 2-bit degradation is a property of model--benchmark pairs, not benchmarks alone. The same quantization can be safe for a strong model on one task, process-degrading for another, and collapse-inducing for a smaller model.
\end{tcolorbox}

\section{Regimes of 2-bit Degradation}
\label{sec:regimes}
The previous section shows that 2-bit quantization affects not only final accuracy, but also the structure of reasoning traces. Importantly, this behavior is not determined by the benchmark alone: the same task can remain stable for a larger model while becoming unstable for a smaller one. We therefore analyze model--benchmark pairs, rather than benchmarks in isolation.

To summarize this behavior, we use two full-budget baseline signals for each pair: the signed accuracy change
\begin{equation}
\label{eq:drop}
\Delta_{2bit}=\mathrm{Acc}_{2bit}-\mathrm{Acc}_{FP16},
\end{equation}
and the think-closed rate under 2-bit inference. Accuracy tells us how much quality is lost, while think-closed indicates whether the low-bit model still produces a well-formed reasoning trace. The baseline columns of Table~\ref{tab:2bit_regimes_and_variants} report these signals and group model--benchmark pairs by the corresponding failure pattern.

The final partition has a simple interpretation:
\begin{itemize}[nosep,leftmargin=*]
    \item \textbf{Low-bit--stable:} 2-bit inference stays close to FP16 and closes traces normally. Qwen3-32B on ARC and GSM8K belongs to this regime because these tasks are already near-saturated for the larger model, so extreme quantization does not visibly damage execution.

    \item \textbf{Precision-sensitive:} accuracy drops, but traces remain well-formed. Qwen3-32B on PIQA, StrategyQA, and WinoGrande fits this pattern: the model usually completes the trace, but becomes less reliable in applying the facts, relations, or commonsense cues needed to reach the correct answer.

    \item \textbf{Process-degraded:} the low-bit model remains partially useful, but generation becomes unstable. This includes Qwen3-32B on MATH-500/GPQA-Diamond and several Qwen3-8B pairs, where the model often fails to maintain or finish the reasoning trajectory.

    \item \textbf{Collapse / fallback-needed:} 2-bit is not enough for direct execution, as in Qwen3-32B on AIME2026 and Qwen3-8B on MATH-500/GPQA-Diamond. It requires fallback, higher precision, or avoiding 2-bit execution.
\end{itemize}

The distinction between precision-sensitive and process-degraded regimes also validates the taxonomy: different failure modes should respond to different interventions. In the \textit{precision-sensitive} regime, traces close normally, so errors are likely local---wrong facts, relations, or commonsense cues. In the \textit{process-degraded} regime, the trace itself becomes unstable, suggesting that the low-bit model struggles to maintain or finish a trajectory.

This motivates a planner--executor test. Following prior work where a stronger model plans for a cheaper executor~\citep{cope,discipl}, we let FP16 write a short plan and condition the 2-bit model on it. If our taxonomy is meaningful, planning should help mainly in the process-degraded regime, where an external plan can stabilize the trajectory, and have little effect when traces are already well-formed.

The $+P$ column in Table~\ref{tab:2bit_regimes_and_variants} matches this prediction. Planning is nearly neutral in the low-bit--stable and precision-sensitive regimes, but gives the largest average gain in the process-degraded regime. This suggests that FP16 planning primarily stabilizes disrupted reasoning, rather than repairing local factual or commonsense errors. In the collapse regime, planning improves accuracy but remains far below FP16, so these cases still require fallback, higher precision, or avoiding 2-bit execution.

One explanation is that the 2-bit model inherits the same underlying knowledge from the FP16 checkpoint, but is worse at applying it coherently. An FP16 plan can therefore guide the low-bit model along a valid solution path, which explains why planning helps selectively in process-degraded cases.

\begin{tcolorbox}[
    colback=takeawaybg,
    colframe=takeawayborder,
    boxrule=0.5pt,
    arc=2pt,
    left=7pt,
    right=7pt,
    top=6pt,
    bottom=6pt,
    boxsep=0pt,
    enhanced
]

\textbf{Takeaway.}
Accuracy drop and think-closed rate separate 2-bit degradation into interpretable regimes. This regime view is predictive: FP16 planning gives little benefit when traces already close normally, but helps most when quantization destabilizes the reasoning process.
\end{tcolorbox}

\renewcommand{\accgain}[2]{#1\,{\scriptsize(#2)}}
\renewcommand{\avgaccgain}[2]{#1\,{\scriptsize(#2)}}  
\newcommand{\baccgain}[2]{\textbf{#1}\,{\scriptsize(#2)}} 

\begin{table*}[t]
\centering
\small
\caption{
Model--benchmark regimes and selective-precision variants under 2-bit quantization.
Drop is computed as $\mathrm{Acc}_{2bit} - \mathrm{Acc}_{FP16}$ (see \cref{eq:drop}).
Drop and $\mathrm{TC}_{2b}$ are the grouping metrics used to define the regimes.
$+P$, $+L$, and $+P+L$ denote FP16 plan, loop rescue, and their combination.
Numbers in parentheses are gains over direct 2-bit execution.
\textbf{Bold} indicates the best result among 2bit, $+P$, $+L$, and $+P+L$ in each row.
Average rows report macro-averages over pairs within each regime.
}
\label{tab:2bit_regimes_and_variants}
\vspace{-2mm}
\begingroup
\renewcommand{\arraystretch}{0.84}
\setlength{\tabcolsep}{3pt}
\begin{tabular}{
llrrrr
>{\columncolor{interventionbg}}c
>{\columncolor{interventionbg}}c
>{\columncolor{interventionbg}}c
}
\toprule
& & \multicolumn{2}{c}{\textbf{Baselines}}
& \multicolumn{2}{c}{\textbf{Metrics}}
& \multicolumn{3}{c}{\cellcolor{interventionbg}\textbf{Selective variants}} \\
\cmidrule(lr){3-4}
\cmidrule(lr){5-6}
\cmidrule(lr){7-9}
Model & Benchmark
& FP16 & 2bit & Drop & $\mathrm{TC}_{2b}$
& $+P$ & $+L$ & $+P+L$ \\
\midrule

\multicolumn{9}{l}{\textbf{I. Low-bit--stable}} \\
\midrule
Qwen3-32B & ARC-Easy
& 94.78 & 94.36 &  -0.42 & 99.92
& \baccgain{94.82}{+0.46} & \accgain{94.19}{-0.17} & \accgain{94.61}{+0.25} \\

Qwen3-32B & ARC-Challenge
& 95.22 & 93.43 &  -1.79 & 99.83
& \baccgain{93.77}{+0.34} & \accgain{93.60}{+0.17} & \accgain{93.69}{+0.26} \\

Qwen3-32B & GSM8K
& 95.83 & 93.33 &  -2.50 & 99.77
& \accgain{92.49}{-0.84} & \baccgain{93.48}{+0.15} & \accgain{92.65}{-0.68} \\

\cmidrule(lr){1-9}
\rowcolor{avgrowbg}
\multicolumn{2}{l}{\cellcolor{avgrowbg}\textbf{Average}}
& 95.28 & 93.71 & -1.57 & 99.84
& \avgaccgain{93.69}{-0.01}
& \baccgain{93.76}{+0.05}
& \avgaccgain{93.65}{-0.06} \\

\midrule
\multicolumn{9}{l}{\textbf{II. Precision-sensitive}} \\
\midrule
Qwen3-32B & PIQA
& 93.74 & 88.25 &  -5.49 & 99.67
& \accgain{88.52}{+0.27} & \accgain{88.47}{+0.22} & \baccgain{89.01}{+0.76} \\

Qwen3-32B & StrategyQA
& 78.89 & 73.36 &  -5.53 & 99.85
& \accgain{73.22}{-0.14} & \baccgain{73.51}{+0.15} & \accgain{72.93}{-0.43} \\

Qwen3-32B & WinoGrande
& 88.48 & 77.51 & -10.97 & 98.03
& \accgain{78.06}{+0.55} & \accgain{78.45}{+0.94} & \baccgain{78.53}{+1.02} \\

\cmidrule(lr){1-9}
\rowcolor{avgrowbg}
\multicolumn{2}{l}{\cellcolor{avgrowbg}\textbf{Average}}
& 87.04 & 79.71 & -7.33 & 99.18
& \avgaccgain{79.93}{+0.23}
& \avgaccgain{80.14}{+0.44}
& \baccgain{80.16}{+0.45} \\

\midrule
\multicolumn{9}{l}{\textbf{III. Process-degraded}} \\
\midrule
Qwen3-32B & MATH-500
& 93.00 & 65.00 & -28.00 & 84.20
& \accgain{80.60}{+15.60} & \accgain{73.20}{+8.20} & \baccgain{87.20}{+22.20} \\

Qwen3-32B & GPQA-Diamond
& 66.16 & 42.93 & -23.23 & 80.81
& \accgain{48.22}{+5.29} & \baccgain{52.53}{+9.60} & \accgain{51.78}{+8.85} \\

Qwen3-8B & ARC-Easy
& 94.32 & 85.14 &  -9.18 & 85.77
& \baccgain{89.31}{+4.17} & \accgain{86.08}{+0.94} & \accgain{88.34}{+3.20} \\

Qwen3-8B & ARC-Challenge
& 93.09 & 75.60 & -17.49 & 76.02
& \baccgain{81.31}{+5.71} & \accgain{76.44}{+0.84} & \accgain{80.03}{+4.43} \\

Qwen3-8B & GSM8K
& 96.13 & 52.69 & -43.44 & 27.14
& \accgain{65.73}{+13.04} & \accgain{74.45}{+21.76} & \baccgain{80.06}{+27.37} \\

Qwen3-8B & PIQA
& 88.36 & 67.63 & -20.73 & 53.10
& \accgain{71.80}{+4.17} & \accgain{67.79}{+0.16} & \baccgain{72.60}{+4.97} \\

Qwen3-8B & StrategyQA
& 73.65 & 62.74 & -10.91 & 81.08
& \accgain{64.05}{+1.31} & \accgain{67.54}{+4.80} & \baccgain{68.56}{+5.82} \\

Qwen3-8B & WinoGrande
& 81.29 & 52.41 & -28.88 & 31.97
& \accgain{54.60}{+2.19} & \baccgain{58.33}{+5.92} & \baccgain{58.33}{+5.92} \\

\cmidrule(lr){1-9}
\rowcolor{avgrowbg}
\multicolumn{2}{l}{\cellcolor{avgrowbg}\textbf{Average}}
& 85.75 & 63.02 & -22.73 & 65.01
& \avgaccgain{69.45}{+6.44}
& \avgaccgain{69.55}{+6.53}
& \baccgain{73.36}{+10.35} \\

\midrule
\multicolumn{9}{l}{\textbf{IV. Collapse / fallback-needed}} \\
\midrule
Qwen3-32B & AIME2026
& 70.00 & 16.67 & -53.33 & 66.67
& \accgain{16.67}{+0.00} & \accgain{36.67}{+20.00} & \baccgain{40.00}{+23.33} \\

Qwen3-8B & MATH-500
& 92.40 & 17.20 & -75.20 & 23.60
& \accgain{26.40}{+9.20} & \accgain{74.20}{+57.00} & \baccgain{76.20}{+59.00} \\

Qwen3-8B & GPQA-Diamond
& 59.09 & 17.17 & -41.92 & 31.31
& \accgain{17.17}{+0.00} & \accgain{47.98}{+30.81} & \baccgain{52.53}{+35.36} \\

\cmidrule(lr){1-9}
\rowcolor{avgrowbg}
\multicolumn{2}{l}{\cellcolor{avgrowbg}\textbf{Average}}
& 73.83 & 17.01 & -56.82 & 40.53
& \avgaccgain{20.08}{+3.07}
& \avgaccgain{52.95}{+35.94}
& \baccgain{56.24}{+39.23} \\

\bottomrule
\end{tabular}
\endgroup
\end{table*}

\section{Loop Rescue for Long-Tail Failures}
\label{sec:loop_rescue}
The regime analysis separates local quality degradation from process-level failures. We now focus on a related practical issue: long-tail generation. As shown in the length analysis, 2-bit inference can lose its practical speed advantage when the model produces long traces, hits the generation budget, or enters repetitive reasoning. In such cases, the model is cheaper per token, but the total generation cost can still erase the expected speedup.

Our loop analysis suggests that repetitions should be treated differently depending on when they occur. Some loops appear after the model has already produced a parseable answer -- these are mostly commitment failures. Other loops appear before any valid answer is produced, suggesting that the 2-bit executor does not reliably solve the 
example.

We therefore use loop detection as a lightweight rescue trigger. During 2-bit generation, we track repeated $n$-grams in the reasoning trace. When a loop is detected, we distinguish two cases:
\begin{itemize}[nosep, leftmargin=*]
    \item \textbf{Commit:} if a parseable final answer appeared before the loop, we return that answer and stop generation.
    \item \textbf{Fallback:} if no parseable answer appeared before the loop,
    we rerun the generation from the beginning in FP16.
\end{itemize}

\paragraph{Result analysis.}
We now check whether loop rescue achieves the two goals it was designed for: recovering accuracy in regimes with unstable traces and reducing the long-tail generation cost that erases the advantage of 2-bit inference.
The $+L$ column in Table~\ref{tab:2bit_regimes_and_variants} shows that rescue is nearly neutral in Regimes I--II, but improves Regimes III--IV, where loops and unfinished traces are common. The $+P+L$ column shows that planning and rescue are complementary: the former stabilizes the trajectory before execution, while the latter handles repetitive failures after they appear.

\newcommand{\lenred}[4]{%
  \shortstack[c]{%
    #1 / #2\\[0.25ex]
    {\scriptsize($\downarrow$#3 / #4)}%
  }%
}

\newcommand{\leninc}[4]{%
  \shortstack[c]{%
    #1 / #2\\[0.25ex]
    {\scriptsize($\downarrow$#3 / $\uparrow$#4)}%
  }%
}
\newcommand{\blenred}[4]{\makecell{\textbf{#1} / \textbf{#2} \\ \scriptsize $-$#3 / $-$#4}}

\begin{table*}[t]
\centering
\caption{
Average reasoning lengths by degradation regime and setup.
Each cell reports mean / median length, macro-averaged over pairs within the regime.
For $+L$ and $+P+L$, the second line shows relative mean / median reduction compared to direct 2-bit inference.
\textbf{Bold} indicates the shortest mean reasoning length among FP16, 2bit, $+P$, $+L$, $+P+L$ in each regime.
}
\label{tab:lengths_by_regime}

\begingroup
\small
\renewcommand{\arraystretch}{0.98}
\setlength{\tabcolsep}{5pt}
\begin{tabular}{lccccc}
\toprule
Regime & FP16 & 2bit & $+P$ & $+L$ & $+P+L$ \\
\midrule
I. Low-bit--stable
& 861 / 689
& 1027 / 492
& 1455 / 552
& \lenred{\textbf{736}}{\textbf{484}}{28.3\%}{1.6\%}
& \leninc{982}{547}{4.4\%}{11.2\%} \\

II. Precision-sensitive
& 396 / 333
& 738 / 283
& 870 / 347
& \lenred{\textbf{373}}{\textbf{282}}{49.5\%}{0.4\%}
& \leninc{477}{343}{35.4\%}{21.2\%} \\

III. Process-degraded
& 1792 / 1380
& 12222 / 11964
& 8615 / 6105
& \lenred{\textbf{949}}{\textbf{754}}{92.2\%}{93.7\%}
& \lenred{1568}{1085}{87.2\%}{90.9\%} \\

IV. Collapse / fallback-needed
& 9111 / 8026
& 24978 / 29164
& 25053 / 30754
& \lenred{4678}{3908}{81.3\%}{86.6\%}
& \lenred{\textbf{4589}}{\textbf{4197}}{81.6\%}{85.6\%} \\
\bottomrule
\end{tabular}
\endgroup
\end{table*}

\newcommand{\bspeedup}[2]{\textbf{#1}\,/\,\textbf{#2}}

\begin{table*}[t]
\centering
\caption{
Weighted quality--speed tradeoff by regime.
Gains are relative to 2-bit; speedups are relative to FP16 and reported as batch-1 / batch-8.
\textbf{Bold} indicates the best Gain and the best Speedup in each regime.
}
\label{tab:weighted_quality_speed_by_regime}

\begingroup
\small
\renewcommand{\arraystretch}{0.95}
\setlength{\tabcolsep}{4pt}
\begin{tabular}{lrrrrrrr}
\toprule
& & \multicolumn{2}{c}{2bit} 
& \multicolumn{2}{c}{$+L$} 
& \multicolumn{2}{c}{$+P+L$} \\
\cmidrule(lr){3-4}
\cmidrule(lr){5-6}
\cmidrule(lr){7-8}
Regime & $N$
& Acc. & Speedup
& Gain & Speedup
& Gain & Speedup \\
\midrule
I. Low-bit--stable
& 4867
& 93.86 & 2.59 / 1.43
& 0.00 & \bspeedup{3.40}{1.93}
& \textbf{+0.00} & 2.22 / 1.36 \\

II. Precision-sensitive
& 3792
& 81.96 & 1.63 / 0.90
& +0.45 & \bspeedup{3.08}{1.74}
& \textbf{+0.63} & 1.93 / 1.25 \\

III. Process-degraded
& 9357
& 67.89 & 0.23 / 0.22
& +5.24 & \bspeedup{2.21}{1.98}
& \textbf{+8.80} & 1.81 / 1.46 \\

IV. Collapse / fallback-needed
& 728
& 17.17 & 0.44 / 0.42
& +48.35 & \bspeedup{1.09}{1.03}
& \textbf{+51.10} & 1.08 / 1.02 \\
\bottomrule
\end{tabular}
\endgroup
\vspace{-4mm}
\end{table*}

Table~\ref{tab:lengths_by_regime} shows that loop rescue primarily controls long-tail cost in the regimes with process failures. In Regimes I--II, the mean length can decrease because rescue trims rare long traces, but the median changes little, confirming that loops are not the typical failure mode there. In contrast, Regimes III--IV show large reductions in both mean and median length. In the process-degraded regime, $+L$ cuts mean / median length by 92.2\% / 93.7\%, and $+P+L$ still cuts them by 87.2\% / 90.9\%. Collapse cases also shorten by over 80\%, but mostly via fallback not reliable execution.

\begin{tcolorbox}[
    colback=takeawaybg,
    colframe=takeawayborder,
    boxrule=0.5pt,
    arc=2pt,
    left=7pt,
    right=7pt,
    top=6pt,
    bottom=6pt,
    boxsep=0pt,
    enhanced
]
\textbf{Takeaway.}
Loop rescue addresses the long-tail failure of 2-bit reasoning: when traces become repetitive, we either commit to an existing answer or rerun in FP16. This yields the largest accuracy and length gains in process-degraded pairs, while collapse pairs still require frequent fallback.
\end{tcolorbox}







\section{Quality--Speed Tradeoff}
\label{sec:quality_speed}

We now turn the length analysis into an end-to-end quality--speed comparison. Runtime depends on both the number of generated tokens and the throughput of the precision mode used to generate them. Throughput measurements for different batch sizes are provided in Appendix~\ref{app:throughput}.

Table~\ref{tab:weighted_quality_speed_by_regime} shows that the speedup story is regime-dependent. In the easy regimes, 2-bit inference is already an efficient path, and loop rescue mostly trims rare long tails. The main win appears in the process-degraded regime: raw 2-bit is slower than FP16 because long traces erase the per-token throughput advantage, while loop rescue restores a clear speedup and improves quality by +5.24 points over direct 2-bit. Adding an FP16 plan improves quality further, to +8.80 points, but reduces speed, giving an explicit quality--speed tradeoff.

This tradeoff is useful in practice. For inputs that are likely easy, the fast low-bit path is preferable; for harder inputs, spending some FP16 computation can improve reliability. This resembles reasoning-effort controls in user-facing systems, where latency and answer quality are traded off depending on task difficulty.

Batch size also matters: weight-only speedups are larger at batch 1, while at batch 8 weight-loading cost is amortized. For serving, the bottleneck shifts toward activation and KV-cache memory. Large speed gains then require not only smaller memory footprints, but also efficient native FP4 computation. While 2-bit is important for studying extreme compression, practical serving may favor stable 4-bit regimes, which we study in Appendix~\ref{app:w4a4}.

\section{Conclusion}
We showed that 2-bit quantization degrades LRMs not only in final-answer accuracy, but in the structure of the reasoning process itself. 
By analyzing full reasoning traces, we identified two dominant failure modes -- path-finding failure and commitment failure -- and introduced a regime-based taxonomy that predicts which interventions are effective for a given model--benchmark pair. 

\section*{Limitations}

Our study is conducted under GPTQ-style weight-only quantization (W2A16, W4A16). While these represent a practically relevant and widely used setting, more advanced quantization methods such as QuIP\# could in principle yield better weight reconstruction at 2-bit and partially alleviate the failure modes we identify. However, their current implementations are largely restricted to older model generations and are not compatible with modern inference accelerators such as vLLM, making large-scale evaluation under realistic serving conditions infeasible at this time. 

Similarly, our experiments are limited to two scales of the Qwen3 family. Extending the study to other recently released reasoning models such as DeepSeek-R1 or Llama-4 is hindered by the same compatibility constraints -- publicly available 2-bit quantizations for these architectures are scarce -- and evaluating larger model scales (e.g., 70B+) is prohibitively expensive under our resource budget, as aggressive quantization experiments require repeated full-budget inference runs across multiple benchmarks. Extending the proposed interventions to both advanced quantization methods and newer model families therefore remains an important direction for future work.
\end{mainpart}

\newpage
\begin{appendixpart}
\tableofcontents
\allowdisplaybreaks
\newpage

\section{Throughput Measurements}
\label{app:throughput}

End-to-end speedups depend not only on generation length, but also on the decoding throughput of each precision mode. Table~\ref{tab:throughput_measurements} reports the throughput measurements used in our speedup estimates. These numbers should not be interpreted as end-to-end speedups by themselves: they only describe token generation speed, while total runtime also depends on how many tokens each method generates and how often FP16 fallback is used.

\begin{table}[ht]
\centering
\small
\caption{
Measured decoding throughput on Qwen3.
Values are reported in generated tokens per second.
}
\label{tab:throughput_measurements}

\begingroup
\renewcommand{\arraystretch}{0.95}
\setlength{\tabcolsep}{6pt}
\begin{tabular}{lrrrrrr}
\toprule
& \multicolumn{3}{c}{Batch 1} 
& \multicolumn{3}{c}{Batch 8} \\
\cmidrule(lr){2-4}
\cmidrule(lr){5-7}
Model & FP16 & 2bit & Ratio & FP16 & 2bit & Ratio \\
\midrule
8B  & 79  & 128 & 1.62$\times$ & 433 & 679 & 1.57$\times$ \\
32B & 23  & 70  & 3.04$\times$ & 117 & 197 & 1.68$\times$ \\
\bottomrule
\end{tabular}
\endgroup
\end{table}

The throughput advantage decreases at larger batch sizes, especially for weight-only quantization, because batching reduces the relative importance of memory bandwidth. This is why we report speedups separately for batch 1 and batch 8 in the main text.

\section{Comparing Loop Detection Criteria}
\label{app:loop_detection_ablation}

Loop rescue depends on how we detect that a reasoning trace has entered an unproductive mode. Our default detector is repetition-based: it triggers when, within the last $t=1024$ generated tokens, some $n$-gram appears at least $m=4$ times. As in the trace-level analysis, we use $n=20$.

An alternative is uncertainty-based early exit. Methods such as DEER~\citep{yang2025deer} monitor hesitation markers such as \texttt{Wait} or \texttt{Alternatively}, or high-entropy reasoning steps. These are useful general-purpose baselines because they target model uncertainty rather than a specific low-bit pathology.

In our setting, the repetition-based detector better matches the failure modes induced by aggressive quantization, where traces often cycle locally instead of making progress. On Qwen3-32B / MATH-500, the uncertainty-based baseline reaches 69 accuracy points in the $+L$ pipeline, while our detector reaches 73. The gain is modest, but consistent with the fact that our method is tailored to recurrent loop failures in low-bit reasoning traces.

\begin{table}[ht]
\centering
\caption{
Ablation of loop-detection criteria on Qwen3-32B / MATH-500.
We compare direct full-precision and 2-bit baselines with two loop-rescue detectors: a general uncertainty-based detector in the style of DEER, and our repetition-based detector.
Our repetition-based detector gives the best result among the 2-bit methods.
}
\label{tab:loop_detection_ablation}

\begingroup
\renewcommand{\arraystretch}{0.95}
\setlength{\tabcolsep}{6pt}
\begin{tabular}{lr}
\toprule
Setup & Accuracy \\
\midrule
FP16 & 93 \\
\midrule
2bit & 65 \\
Uncertainty-based detector & 69 \\
\textbf{Repetition-based detector (ours)} & \textbf{73} \\
\bottomrule
\end{tabular}
\endgroup
\end{table}

\section{High-Batch Deployment with W4A4 and Low-Precision KV Cache}
\label{app:w4a4}

The main text focuses on extreme 2-bit inference because it exposes the failure modes of low-bit reasoning most clearly. 
However, 2-bit execution is not the only practical route to faster reasoning. 
As discussed in Section~\ref{sec:related}, 4-bit quantization is often much more stable in quality. 
For high-throughput serving, the remaining question is therefore different: if 4-bit weights already preserve reasoning quality, where does the next speedup come from?

At larger batch sizes, weight-only quantization becomes less decisive because the KV cache and activation precision start to dominate memory usage and serving capacity. 
This makes low-precision KV-cache and FP4-style execution a natural deployment direction. 
In this appendix, we report preliminary single-GPU measurements on an NVIDIA B200 to illustrate this effect. 
These results are not intended as a fully optimized serving benchmark; rather, they show that on B200-class hardware, W4A4/NVFP4 execution with low-precision KV cache can be a highly practical regime for large-batch reasoning.

\paragraph{Deployment effect.}

Table~\ref{tab:b200_kv_deployment} shows that at 32K maximum generation length, KV-cache precision has a large impact on the maximum feasible batch size. 
Reducing the KV cache from FP16 to FP8 roughly doubles the number of concurrent sequences, while NVFP4 KV gives another substantial increase. 
This effect is complementary to weight quantization: for Qwen3-32B, combining quantized weights with NVFP4 KV increases the feasible batch size from single digits to more than 60 concurrent sequences; for Qwen3-8B, the same setup reaches around 120 concurrent sequences.

\begin{table*}[t]
\centering
\caption{
B200 deployment measurements at 32K maximum generation length.
Max batch denotes the largest feasible number of concurrent sequences; throughput is measured at batch size 8 in generated tokens per second.
}
\label{tab:b200_kv_deployment}
\begingroup
\renewcommand{\arraystretch}{0.95}
\setlength{\tabcolsep}{5pt}
\begin{tabular}{lrrrrrr}
\toprule
& \multicolumn{3}{c}{Max batch} 
& \multicolumn{3}{c}{Throughput} \\
\cmidrule(lr){2-4}
\cmidrule(lr){5-7}
Setup & FP16 KV & FP8 KV & NVFP4 KV & FP16 KV & FP8 KV & NVFP4 KV \\
\midrule
Qwen3-32B FP16   & 9  & 19 & 39  & 456  & 528  & 523 \\
Qwen3-32B W4A16  & 15 & 30 & 61  & 614  & 740  & 738 \\
Qwen3-8B W4A16   & 29 & 59 & 119 & 1312 & 1634 & 1639 \\
\bottomrule
\end{tabular}
\endgroup
\end{table*}

The throughput gains are smaller than the max-batch gains, but they follow the same trend: lower-precision KV improves serving efficiency. 
This is especially relevant for batched reasoning workloads, where the practical bottleneck is often not the speed of a single sequence, but the number of long-context generations that can be served concurrently. 
Thus, while the main paper studies selective control for fragile 2-bit reasoning, Table~\ref{tab:b200_kv_deployment} shows a complementary deployment message: when B200-class FP4 support is available, 4-bit execution with low-precision KV cache is a strong high-batch serving regime.

\paragraph{Quality.}

The deployment gains above are useful only if FP4-style execution preserves reasoning quality. 
Table~\ref{tab:nvfp4_quality_selected} therefore reports a compact quality comparison between FP16 and NVFP4 setups. 
The FP16 rows use the same macro-averages as the corresponding main-table evaluations; the NVFP4 rows are evaluated over the same benchmark sets for each model.

\begin{table}[t]
\centering
\caption{
Selected quality results for FP16 and NVFP4 setups.
Accuracy is macro-averaged over the same evaluated benchmark set for each model.
}
\label{tab:nvfp4_quality_selected}
\begingroup
\renewcommand{\arraystretch}{0.95}
\setlength{\tabcolsep}{6pt}
\begin{tabular}{lrr}
\toprule
Setup & Accuracy & Think-closed \\
\midrule
Qwen3-8B FP16    & 84.79 & 100.0 \\
Qwen3-8B NVFP4   & 80.90 & 100.0 \\
Qwen3-32B FP16   & 86.23 & 100.0 \\
Qwen3-32B NVFP4  & 85.29 & 99.9 \\
\bottomrule
\end{tabular}
\endgroup
\end{table}

Table~\ref{tab:nvfp4_quality_selected} suggests that NVFP4 preserves reasoning quality much better than extreme 2-bit execution, especially for Qwen3-32B. 
The remaining degradation is concentrated on harder benchmarks such as AIME and GPQA-Diamond, so we do not treat FP4-style inference as universally lossless. 
Nevertheless, these results make W4A4/NVFP4 with low-precision KV cache a practically attractive option for high-throughput deployment: if the target hardware supports efficient FP4 execution, it can provide large serving-capacity gains while avoiding many of the process-level failures observed under 2-bit inference.

\section{Effect of Sampling Temperature}
\label{app:temperature}

Temperature is an important but under-standardized evaluation choice for reasoning models. In the main experiments, we use the default Qwen-style setting, $t=0.6$. However, different papers often use different decoding temperatures, and this can substantially change both quality and process-level metrics. This makes direct comparisons between quantized reasoning results harder unless the decoding setup is reported explicitly.

\begin{table}[ht]
\centering
\caption{
Temperature ablation for GPTQ-2bit reasoning.
Loop and HitMax denote the fraction of generations with detected loops and maximum-budget termination.
}
\label{tab:temperature_ablation}
\begingroup
\renewcommand{\arraystretch}{0.95}
\setlength{\tabcolsep}{5pt}
\begin{tabular}{lrrrr}
\toprule
Run & Acc. & Loop & HitMax & Avg. len \\
\midrule
32B, $t=0.0$ & 60.6 & 28.8 & 26.2 & 10248 \\
32B, $t=0.6$ & 65.0 & 18.0 & 13.2 & 7325 \\
\midrule
8B, $t=0.0$ & 16.0 & 94.6 & 98.0 & 30140 \\
8B, $t=0.6$ & 17.2 & 90.2 & 93.4 & 28814 \\
\bottomrule
\end{tabular}
\endgroup
\end{table}

Table~\ref{tab:temperature_ablation} shows a small ablation on 2-bit reasoning. Changing temperature affects not only accuracy, but also loop rate, maximum-budget hits, and average generation length. For Qwen3-32B GPTQ-2bit, moving from greedy decoding to $t=0.6$ improves accuracy from 60.6\% to 65.0\%, while reducing loops, budget hits, and average length. For Qwen3-8B GPTQ-2bit, however, both settings remain collapsed: accuracy stays near 16--17\%, and almost all generations loop or hit the maximum budget. Thus, temperature can improve borderline regimes, but it does not by itself fix severe low-bit collapse.

Overall, temperature affects not only accuracy, but also looping, budget hits, and generation length. We therefore keep it fixed in the main experiments and treat it as part of the evaluation protocol, not as an incidental implementation detail.

\section{Quality--Speed Pareto Frontier}
\label{app:quality_speed_pareto}

As shown in Figure~\ref{fig:qwen3_32b_quality_speed_pareto_bs1}, although direct 2-bit inference achieves high token throughput, it does not lie on the Pareto frontier because loop-related failures often lead to much longer generations. As a result, end-to-end problem solving remains slow despite the higher raw decoding speed. Our selective 2-bit pipeline mitigates this effect, improving both quality and end-to-end speed and thereby placing the model on the Pareto frontier.

\begin{figure}[ht]
\centering
\includegraphics[width=0.5\columnwidth]{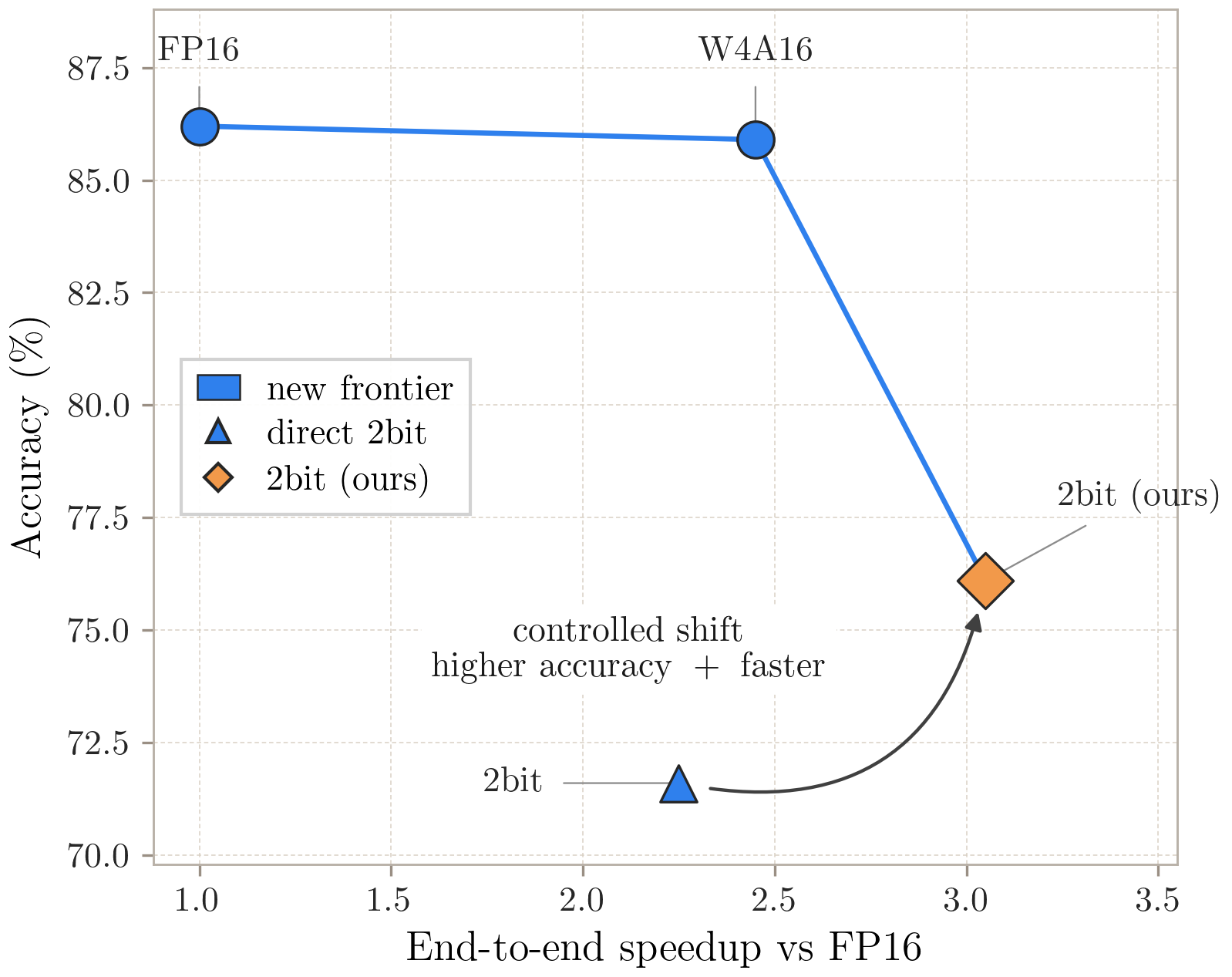}
\caption{
Quality--speed Pareto frontier for Qwen3-32B at batch size 1.
}
\label{fig:qwen3_32b_quality_speed_pareto_bs1}
\end{figure}

\section{Experimental Reproducibility Checklist}
\label{app:repro}

\subsection{Hardware}
\label{app:repro:hardware}

Experiments were run on an NVIDIA A100 GPU unless otherwise stated.
Experiments involving NVFP4 and MoE inference were run on NVIDIA B200 GPUs. Full evaluation time for 2 models of about $100$ GPU-hours in our setup.
Throughput measurements used in Appendix~\ref{app:throughput} and Section~\ref{sec:quality_speed} were collected in the same experimental setup.

\subsection{Evaluation}
\label{app:repro:evaluation}

The main evaluation setup is described in Section~\ref{sec:setup}.
Unless otherwise stated, reported accuracy, reasoning length, and trace-level metrics are from a single quantization/evaluation run under the specified recipe.
All precision regimes use the same prompt format, decoding configuration, and answer parser whenever possible.

Answers are extracted with the same task-specific parser across precision regimes.
For mathematical tasks (AIME, MATH-500, GSM8K), the parser first extracts the final answer from the last \texttt{\textbackslash boxed\{\}} span when present and otherwise falls back to the final numeric answer; for AIME, the parsed answer is normalized to an integer in the target range.
For multiple-choice benchmarks (GPQA, ARC-Easy, ARC-Challenge), we parse the final answer letter, for PIQA and WinoGrande the final choice index (\texttt{1}/\texttt{2}), and for StrategyQA the final \texttt{yes}/\texttt{no} token.
If no valid answer can be parsed, the example is counted as incorrect.

\subsection{Inference and intervention hyperparameters}
\label{app:repro:inference_hparams}

The inference setup, generation budgets, trace-level metrics, degradation regimes, and selective-precision interventions are described in Sections~\ref{sec:setup}--\ref{sec:quality_speed}.
Unless stated otherwise, all main experiments use a 32K generation budget and temperature $0.6$; the budget ablation in Section~\ref{sec:budget} additionally evaluates 4K and 8K limits.

We keep decoding parameters, prompt format, and answer parsing fixed across precision regimes whenever possible.
Other decoding parameters were top-$p=0.95$, top-$k=20$, and random seed $0$.

We used quantized Qwen3 checkpoints based on the Qwen3 model family from the kaitchup Hugging Face collection~\citep{qwen3_hf_collection}.

For loop rescue, we use the repetition-based detector from Section~\ref{sec:loop_rescue}: within a window of $t=1024$ generated tokens, a loop is detected when some $n=20$-gram appears at least $m=4$ times.

\subsection{Code availability}
\label{sec:code}

The repository includes scripts and configurations for reproducing the main experiments (see link in Abstract). In particular, it includes: FP16, W4A16, and W2A16 inference scripts; GPTQ-style quantization configurations; benchmark evaluation and answer parsing; trace-level metric computation; loop detection; loop rescue; FP16 planning; the combined +P+L pipeline; and throughput-based speed estimation.

We also include scripts/configurations to reproduce the accuracy, trace-level, length, and speed results reported in Sections~\ref{sec:setup}--\ref{sec:quality_speed} and Appendices~\ref{app:throughput}--\ref{app:loop_detection_ablation}.
All prompts, decoding parameters, benchmark-specific parsing rules, and intervention settings are provided in the repository.

\subsection{Artifacts and licenses}
\label{app:artifacts_licenses}

We use publicly released Qwen3 checkpoints~\citep{qwen3} and quantized versions ~\citep{qwen3_hf_collection} under their original model licenses and access terms.
We evaluate on publicly available benchmark datasets: AIME 2026~\citep{aime2026}, GPQA-Diamond~\citep{rein2023gpqa}, MATH-500~\citep{hendrycks2021math}, GSM8K~\citep{cobbe2021gsm8k}, StrategyQA~\citep{geva2021strategyqa}, WinoGrande~\citep{sakaguchi2020winogrande}, ARC-Easy/ARC-Challenge~\citep{clark2018arc}, and PIQA~\citep{bisk2020piqa}.
We follow the corresponding model and dataset licenses.

We release anonymized software for reproducing the experiments.
We do not release quantized checkpoints; model checkpoints remain available only through the corresponding original model releases and are subject to their original licenses and access terms.
We do not redistribute benchmark data or calibration text.
The released software follows the upstream licenses stated in the repository README.

\end{appendixpart}
\end{document}